\pdfoutput=1

\documentclass[11pt]{article}

\usepackage[]{acl}

\usepackage{times}
\usepackage{latexsym}

\usepackage[T1]{fontenc}

\usepackage[utf8]{inputenc}

\usepackage{microtype}
\usepackage{graphicx}
\usepackage{caption}
\usepackage{subcaption}
\usepackage{booktabs}
\usepackage{tabularx}
\usepackage{multicol}
\usepackage{multirow}

\title{Comparing Specialised Small and General Large Language Models on Text Classification: 100 Labelled Samples to Achieve Break-Even Performance}

\author{Branislav Pecher$^\spadesuit$$^\dagger$, Ivan Srba$^\dagger$, Maria Bielikova$^\dagger$ \\
$^\spadesuit$ Faculty of Information Technology, Brno University of Technology, Brno, Czechia\\
$^\dagger$ Kempelen Institute of Intelligent Technologies, Bratislava, Slovakia\\
\texttt{\{branislav.pecher, ivan.srba, maria.bielikova\}}@kinit.sk}

\begin{document}
\maketitle
\begin{abstract}
When solving NLP tasks with limited labelled data, researchers typically either use a general large language model without further update, or use a small number of labelled samples to tune a specialised smaller model. In this work, we answer an important question -- how many labelled samples are required for the specialised small models to outperform general large models, while taking the performance variance into consideration. By observing the behaviour of fine-tuning, instruction-tuning, prompting and in-context learning on 8 language models, we identify such performance break-even points across 8 representative text classification tasks of varying characteristics. We show that the specialised models often need only few samples (on average $100$) to be on par or better than the general ones. At the same time, the number of required labels strongly depends on the dataset or task characteristics, with fine-tuning on binary datasets requiring significantly more samples. When performance variance is taken into consideration, the number of required labels increases on average by $100 - 200\%$. Finally, larger models do not consistently lead to better performance and lower variance, with 4-bit quantisation having negligible impact.
\end{abstract}

\section{Introduction}

With the introduction of the GPT-3 model~\cite{brown2020language}, large language models have been shown to be an effective generalist models for learning with limited labelled data. They are able to perform well across many NLP tasks with no (using prompting) or only few (using in-context learning) labelled samples and without any parameter update~\cite{qin2023chatgpt, sun2023pushing, liu2023pretrain-prompt-survey}. This performance is achieved by conditioning the model on an appropriate text input (prompt) containing instructions for the given task, a test sample for which to generate output and optionally a set of few in-context examples showcasing the task~\cite{sun2023pushing, dong2022survey}. Many enhancements were proposed to improve the overall few-shot behaviour. For example prompt-tuning and automatic prompt-engineering, where the effective prompts are designed automatically, or instruction-tuning, where language models are tuned to better follow the task instructions~\cite{gao-etal-2021-making, logan-iv-etal-2022-cutting}.

\begin{figure}[t]
    \centering
    \includegraphics[width=\linewidth]{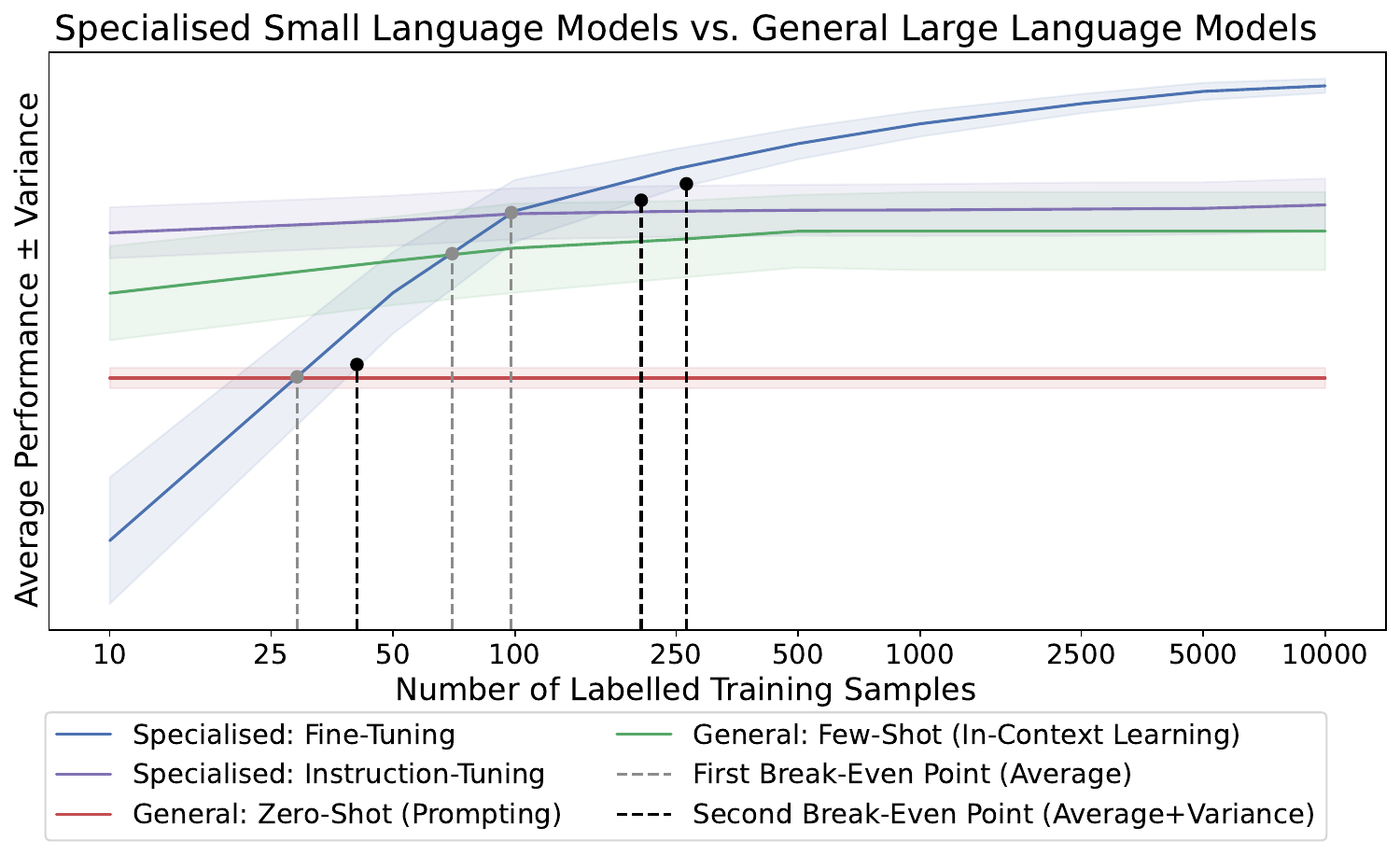}
    \caption{Comparison between the performance of specialised small and general large language models. The break-even points are identified by observing the impact of changing the number of available labelled samples and taking performance variance into consideration. Specialised models outperform general ones with only few labelled samples (up to $100$), with performance variance showing strong impact on the comparison, increasing the number significantly.}
    \label{fig:exp-illustration}
\end{figure}

With enough labelled samples, the \textit{specialised} smaller models, obtained through fine-tuning or instruction tuning, can achieve performance on par or better than the \textit{general} large language models used with prompting or in-context learning without further parameter update~\cite{schick-schutze-2021-just, qin2023chatgpt}. From practical perspective (e.g., to decide how many labelled samples are needed or to choose the best approach when the number of labels is fixed), it is valuable to know (at least approximately) the number of labelled samples needed to achieve such superior performance, i.e., to identify performance break-even points.

So far, most of the studies comparing the performance of specialised and general language models from the data-efficiency perspective have been done using different (often non-representative and incomplete) settings and methodologies, leading to divergent findings~\citep{pecher2023effects}. The specialised language models are tuned either on the \textit{whole labelled set} (where they achieve higher performance than general language models)~\cite{schick-schutze-2021-just, qin2023chatgpt}, or using the same set of \textit{few samples} (e.g., $4-32$) used for general models (where they achieve significantly worse performance than general models)~\cite{ma-etal-2023-large}. Only few works study the behaviour of models between the few samples and the whole set, and observe the performance break-even points where specialised language models outperform general ones. Moreover, they often focus only on specific approaches or ignore the performance variance (caused by various sources of randomness, such as random seeds)~\cite{le-scao-rush-2021-many, hongjin2022selective, lehman2023we, gupta2023instruction, pecher2023effects}.

In addition, almost no work considers the compute-efficiency of the different approaches in the comparisons, even though it is an important aspect when the approaches achieve comparable results~\citep{mosbach-etal-2023-shot}. Although some works investigate the impact of quantisation on model performance~\citep{liu-etal-2024-emergent, jin2024comprehensive}, the impact on the performance variance (i.e., sensitivity) or the overall comparison is not explored.

Our goal is to remedy these shortcomings and answer the following question: \emph{"How many labelled samples do the \textbf{specialised smaller models} need to outperform more \textbf{general larger models}?"} To achieve this, we investigate and observe how the performance of different approaches changes when increasing the number of labelled training samples (\textit{data-efficiency}). We identify the break-even points between performance of specialised models and more general models of different sizes (\textit{compute-efficiency}), while taking the performance variance into consideration (\textit{sensitivity to randomness}). The aggregated results are presented in Figure~\ref{fig:exp-illustration}. Our main contributions and findings are\footnote{To support replicability and extension of our results, we openly publish the source code of our experiments at \url{https://github.com/kinit-sk/L3D-sensitivity-investigation}}:
\begin{itemize}
    \item We perform a comprehensive and fair investigation of the impact of increasing the number of labelled training samples on performance and its variance of fine-tuning, prompting, in-context learning and instruction-tuning approaches across 8 language models and 8 representative text classification datasets of various characteristics, such as number of classes, text length or task type.
    \item By identifying break-even points between the specialised and general models, we find that the smaller specialised models (obtained through fine-tuning or instruction-tuning), require only a small number of labelled training examples ($100$ on average) to achieve performance on par or even better than general large language models used in zero/few-shot settings. In addition, we find significant impact of performance variance, especially originating from in-context learning or fine-tuning on few samples, increasing the number of required labelled samples by up to $500 - 3000\%$, with average increase of $100\%-200\%$.
    \item Based on further analysis we find following key insights: 1) the required number of labelled samples is dependent on the dataset characteristics, with binary datasets and datasets that require better language understanding (e.g., question answering) requiring more labelled samples; 2) larger models do not necessarily lead to better model performance, data-efficiency or lower performance variance for general models; and 3) 4-bit quantisation has negligible impact on the overall behaviour of general models.
    \item Based on the observed findings, we provide practical recommendations on how to choose the best approach, model or to decide how many additional data to label considering the available annotation and computation budget. 
\end{itemize}

\section{Related Work}

Extensive focus is dedicated to the evaluation and comparison between different data efficient approaches utilising large language models, such as prompting, in-context learning, fine-tuning, prompt-tuning or prompt-based/instruction tuning on text classification tasks~\cite{dong2022survey, liu2023pretrain-prompt-survey}. 
In some comparisons, the performance of specialised small models is compared with significantly larger general models~\cite{schick-schutze-2021-just, ma-etal-2023-large, liu2022few, lehman2023we, hongjin2022selective}, while in others the setting is kept as similar as possible (i.e., comparing specialised and general models of the same sizes)~\cite{mosbach-etal-2023-shot, qin2023chatgpt, logan-iv-etal-2022-cutting, gao-etal-2021-making, le-scao-rush-2021-many, gupta2023instruction}. 

Majority of the comparisons focus on comparing fine-tuned models with in-context learning~\cite{ma-etal-2023-large, hongjin2022selective, lehman2023we, qin2023chatgpt}, or on comparing prompt-based/instruction-tuned models on the specific tasks with their general counterparts~\cite{liu2022few, mosbach-etal-2023-shot, schick-schutze-2021-just}. Only few papers compare fine-tuning with instruction-tuning~\cite{gupta2023instruction, le-scao-rush-2021-many} or multiple approaches at the same time (i.e., fine-tuning, instruction-tuning, prompting and in-context learning at the same time)~\cite{logan-iv-etal-2022-cutting, gao-etal-2021-making}.

If enough labelled samples are used, the smaller specialised models can achieve performance on par with, or in many cases even better than the performance of the larger general models~\cite{schick-schutze-2021-just, qin2023chatgpt, lehman2023we}. At the same time, in the extremely limited settings, where the models are fine-tuned using the same low number of samples as it is used for in-context learning, the general large language models excel over the small specialised models~\cite{ma-etal-2023-large, hongjin2022selective}. Other papers study the impact of varying the training data sizes on the performance of specialised models and their comparison with general models. However, they often either perform a comprehensive comparison across multiple approaches, but focus only on few samples and vary the data sizes only to up 32 samples~\cite{logan-iv-etal-2022-cutting, gao-etal-2021-making}, or focus on small subset of approaches while using a larger part or the whole dataset, but often in specific domains~\cite{le-scao-rush-2021-many, gupta2023instruction, hongjin2022selective, lehman2023we, ma-etal-2023-large}. Only few papers consider (to a certain extent) effects of different systematic choices (e.g., number of samples, or prompt format) or sensitivity to sources of randomness (e.g., choice or order of samples) on comparison~\citep{ma-etal-2023-large, pecher2024sensitivity, sclar_quantifying_2023, weber-etal-2023-mind}.

Compared to these works, we focus on more comprehensive comparisons across: 1) full training dataset, increasing the size from 10 samples to full dataset in exponential fashion; 2) multiple approaches for obtaining specialised models (fine-tuning, prompt-based/instruction-tuning) and for using the general models (prompting, in-context learning); 3) multiple runs to carefully take into consideration the sensitivity of different approaches and the performance variance this sensitivity introduces; and 4) models of different sizes.

\section{Comparison Methodology: Identifying Performance Break-Even Points}

Our main focus is on comparing the specialised and general models from the perspective of their data- and compute-efficiency and sensitivity to the effects of randomness, as illustrated by Figure~\ref{fig:exp-illustration}. 

\paragraph{Data-efficiency.} To compare the approaches from the perspective of the data-efficiency, we observe and compare how their performance changes when increasing the number of available labelled samples from low number of samples up until the full dataset is used (where each approach is presented with the same set of samples). As we change the number of available samples, we identify the first break-even point between the performance of the specialised and general models. The first break-even point (average performance) specifies the point after which the performance of the specialised models is better on average, but may still be lower in many cases due to the performance variance and the randomness sensitivity, such as when the randomness is not sufficiently addressed (low number of runs is used or the runs are cherry-picked). At each point, we report the mean F1 macro and standard deviation.

\paragraph{Compute-efficiency.} We compare models of different sizes (e.g., using significantly smaller specialised models) while specifically focusing on models that require a comparable amount of computation. For example, full training and evaluation of the smallest model should require the same amount of compute as using the smallest LLM through prompting. More information are provided in Appendix~\ref{app:computation-flops}. In addition, for selected general models, we explore the impact of quantisation on the performance and its variance.

\paragraph{Sensitivity to the effects of randomness.} To further explore the impact of the sensitivity to the effects of randomness on the comparison, we identify the second break-even points (average+variance). This break-even point denotes the "worst-case size" or the point after which the specialised models show better performance even when the variance is taken into consideration. In other words, there is low probability that the performance of the specialised models will be lower even when using low number of runs or cherry-picking them. To identify this break-even point, we repeat the training and evaluation multiple times and then compare the worst case performance of the better performing model (obtained by subtracting the standard deviation from the mean) with the best case performance of the worse performing model (obtained by adding the standard deviation to the mean). Each repeat uses the same fixed random seed for all models to guarantee deterministic and replicable results. The random seed covers all sources of randomness~\citep{gundersen2022sources_of_irreproducibility, pecher2023effects} -- the split of data, selection of labelled samples, initialisation of models, order of data in training, non-deterministic operations (e.g., dropout), and selection of samples for in-context learning.

\paragraph{Datasets.} The investigation covers 8 classification datasets composed of tasks with different number of classes and characteristics. We focus on 4 binary datasets from GLUE~\cite{wang-etal-2018-glue} and SuperGLUE~\cite{wang2019superglue} benchmarks: \textit{SST2}~\cite{socher-etal-2013-recursive} for sentiment classification, \textit{MRPC}~\cite{dolan-brockett-2005-automatically} for determining semantic equivalence relationship between two sentences, \textit{CoLA}~\cite{warstadt-etal-2019-neural} for determining the grammatical acceptability of a sentence, and \textit{BoolQ}~\cite{clark-etal-2019-boolq} for question answering. In addition, we use 4 multi-class text datasets: \textit{AG News}~\cite{zhang2015agnews} for news classification (4 classes), \textit{TREC}~\cite{voorhees2000trec} for question classification (6 classes), \textit{SNIPS}~\cite{coucke2018snips} for intent classification (7 classes) and \textit{DB Pedia}~\cite{lehmann2015dbpedia} for topic classification (14 classes).

\paragraph{Approaches and models.} We investigate the impact of training dataset size across a set of currently popular approaches for dealing with limited labelled data in NLP: 1) \emph{fine-tuning}; 2) \emph{instruction-tuning}; 3) \emph{prompting}; and 4) \emph{in-context learning}. For fine-tuning, we use the \textbf{BERT}~\cite{devlin2019bert} and \textbf{RoBERTa}~\cite{liu2019roberta} base models. For prompting and in-context learning the \textbf{Flan-T5}~\cite{chung2024scaling} base, \textbf{GPT-4} (4o-mini-2024-07-18 version)~\cite{brown2020language, ouyang2022training}, \textbf{LLaMA-2}~\cite{touvron2023llama} 13B chat optimised model, \textbf{LLaMA-3}~\citep{dubey2024llama} 8B instruction optimised model, \textbf{Mistral-7B}~\cite{jiang2023mistral} and \textbf{Zephyr-7B}~\cite{tunstall2023zephyr} models. For instruction-tuning we use the \textbf{Flan-T5} (full instruction-tuning), \textbf{Mistral-7B} and \textbf{Zephyr-7B} models (both instruction-tuned using LoRA~\cite{hu2021lora}). Due to the significant costs, we use 4-bit quantisation for the LLaMA-2 model. In case of Mistral-7B and Zephyr-7B, we use both the full-precision and the 4-bit quantised versions, reporting only the best performance.

\paragraph{Experimental setup.} Each experiment is repeated 100 times for fine-tuning and 20 times for the remaining approaches (due to the significant costs of inference or training of the larger language models). 
For in-context learning, we use as many samples per class as the context-size of the models allow. As increasing this number after a certain point results in degraded performance, we always report the best performance achieved up until the dataset subset.
For further details see Appendix~\ref{app:experimental-setting-details}.

\subsection{Comparison of Specialised and General Models}
\label{sec:first-bep}

\begin{figure*}[tbh]
    \centering
    \includegraphics[width=0.99\textwidth]{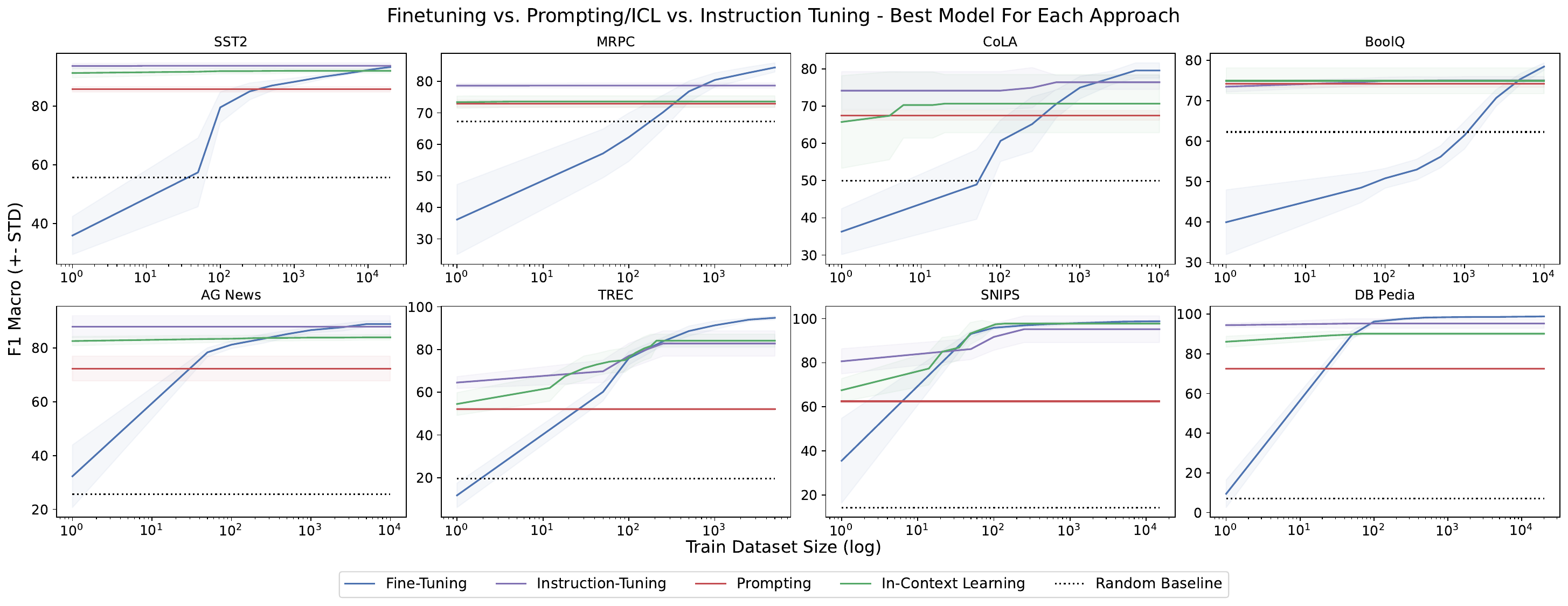}
    \caption{The impact of varying size of available labelled samples (in logarithmic scale) on the performance of fine-tuning, prompting, in-context learning and instruction-tuning approaches, reported using F1 macro and its standard deviation. For each approach, we select only the best performing model. We can observe that \textbf{specialised models can often outperform general models with only a relatively small number of labelled samples ($10 - 1000$)}.}
    \label{fig:best_approach}
\end{figure*}

In this section, our goal is to answer the following main research question: \textit{\textbf{RQ1:} How do the specialised models compare to general models on average as the number of labelled samples increases?} To achieve this, we identify the first break-even point, i.e., the point after which the specialised models outperform the general ones on average. As such, it will allow us to draw findings about the data efficiency and compute efficiency that are then used for the recommendations in Section~\ref{section:recommendations}. The aggregated results for all the models and datasets are presented in Figure~\ref{fig:exp-illustration} and for the best model for each approach across datasets in Figure~\ref{fig:best_approach}.

\textbf{Specialised models can outperform the general ones using only a relatively small number of labelled examples.} To outperform zero-shot setting (prompting), fine-tuning approaches require on average between $10 - 500$ labelled samples. In case of few-shot setting (in-context learning), the number of required labelled samples is higher, on average between $100 - 2000$ samples.

\textbf{The instruction-tuned models provide even larger benefit, representing a good balance between the generality of the models and the samples required for specialisation.} To outperform these models, fine-tuning approaches often require a large fraction of the labelled dataset (on average up to $5000$ samples). In addition, we observe a consistent performance of instruction-tuned models, regardless of how many labelled samples are used. The performance achieved by instruction-tuned models with only $10$ labelled samples is close to the one achieved with the full labelled dataset. \textbf{The difference in performance across dataset subsets is mainly the effect of in-context examples and not the training samples.} Furthermore, the \textbf{instruction-tuned models consistently outperform all the general larger counterparts with small number of samples}. As such, instruction-tuned models achieve the almost-best performance with only a fraction of labelled samples required by fine-tuning, making them an ideal specialised models with highest data-efficiency in many cases.

\begin{figure*}[tbh]
    \centering
    \includegraphics[width=0.99\textwidth]{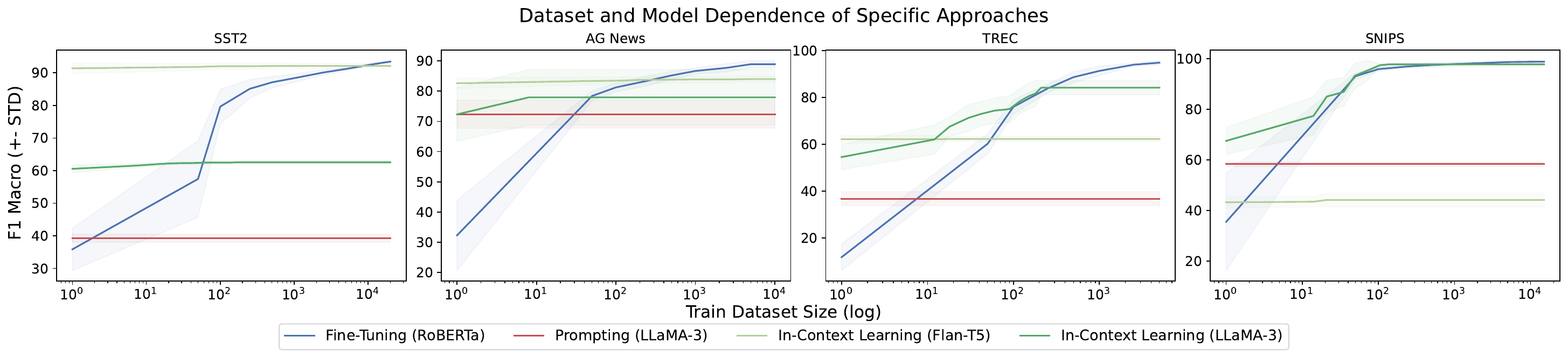}
    \caption{A showcase of the dataset dependence of the break-even points for specific models. The models that perform well on one dataset may perform significantly worse on others, due to the different characteristics, such as the number of classes, sentence length, task type or whether the dataset was used as part of the model pre-training.}
    \label{fig:dataset_dependence}
\end{figure*}

However, \textbf{the comparison between models and the number of required labelled samples is strongly affected by the dataset and task characteristics.} This includes characteristics like 1) how many classes are used (binary vs. multi-class setting); 2) length of sentences (e.g., SST2 vs. CoLA); 3) whether the task requires working with a single sentence or a paired/multiple inputs (e.g., SST2/CoLA vs. MRPC/BoolQ); or 4) the overall type of the task (e.g., simple sentiment classification vs. question answering) that is defined by these characteristics. \textbf{The largest difference can be observed between binary and multi-class datasets}. On multi-class datasets, fine-tuning requires only up to $100-200$ labelled samples to outperform the larger models (even instruction-tuned ones in many cases). On binary datasets the required samples are as high as $4000$ to outperform zero-shot setting (prompting), up to $9000$ to outperform few-shot setting (in-context learning), or up to $20 000$ to outperform the instruction-tuned models. The remaining characteristics have lower impact that is mostly dependent on the model size (more information is provided in Appendix~\ref{sec:model_size_effect}).

An additional significant factor that affects the comparison is \textbf{whether the dataset was used as part of the larger models pre-training.} As illustrated in Figure~\ref{fig:dataset_dependence}, in such a case, the general larger model are significantly more competitive with the specialised models. For example, the in-context learning with LLaMA-3 model on the SNIPS (and TREC to a certain extent) dataset achieves similar performance to the fine-tuning, and increasing the number of samples also leads to increase in performance of both approaches. However, the LLaMA-3 performance is not consistent across datasets as we observe significant drop in performance on the other datasets, such as AG News or SST2. Similar behaviour can be observed for other models and approaches as well, such as Flan-T5 on SST2 or AG News datasets as compared to TREC and SNIPS, or prompting with LLaMA-3. In these cases (when the model is pretrained on the dataset and/or achieves comparable performance), \textbf{it is important to consider the compute-efficiency of the approaches as well}, as it is especially important in the long run (i.e., using the model in practice).

\subsection{Impact of Performance Variance on Comparison}

\begin{table}[tbh]
\tabcolsep=0.15cm
\begin{center}
\resizebox{\linewidth}{!}{
\begin{tabular}{@{}clccc@{}}
\toprule
& & Prompting & In-Context Learning & Instruction-Tuning \\ \midrule
\multirow[c]{8}{*}{\rotatebox{90}{Fine-Tuning}} 
& \textit{SST2}       & $350 ~|~ 900^{+157\%}$      & $9000 ~|~ 20000^{+122\%}$       & $20000 ~|~ NA$ \\
& \textit{MRPC}       & $350 ~|~ 500^{+43\%}$      & $375 ~|~ 700^{+87\%}$       & $750 ~|~ 1750^{+133\%}$ \\
& \textit{CoLA}       & $350 ~|~ 700^{+100\%}$      & $500 ~|~ NA$       & $1800 ~|~ NA$ \\
& \textit{BoolQ}       & $4250 ~|~ 6000^{+41\%}$      & $5000 ~|~ NA$       & $5000 ~|~ 7500^{+50\%}$ \\
& \textit{AGNews}       & $40 ~|~ 60^{+50\%}$      & $300 ~|~ 800^{+167\%}$       & $2500 ~|~ NA$ \\
& \textit{TREC}       & $30 ~|~ 50^{+67\%}$      & $250 ~|~ 600^{+140\%}$       & $200 ~|~ 900^{+350\%}$ \\
& \textit{SNIPS}       & $30 ~|~ 40^{+33\%}$      & $50 ~|~ NA$       & $40 ~|~ NA$ \\
& \textit{DBPedia}       & $40 ~|~ 40^{+0\%}$      & $50 ~|~ 100^{+100\%}$       & $100 ~|~ 500^{+500\%}$ \\
\bottomrule
\end{tabular}}
\end{center}
\caption{Break-even points between the best performing models of different approaches across all investigated datasets. The first break-even point (average performance) and the second one (average and variance) are separated with the "|" symbol, with superscript indicating the percentual increase and "NA" indicating no break-even point exists. We observe a \textbf{significant influence of the performance variance on the number of required labelled samples}.}
\label{tab:break-even-points}
\end{table}

In this section, we answer following research question: \textit{\textbf{RQ2:} How does the variance from repeated runs affect the comparison between specialised and general models?} Our goal is to determine how the number of required samples increases when the performance variance is taken into consideration. To accomplish this, we identify the second break-even point (average+variance) denoting "worst-case sizes" of training sets. Identifying this point will allow us to determine how the sensitivity of the different approaches affect the comparisons. The aggregated results for all the models and datasets are presented in Figure~\ref{fig:exp-illustration} and for the best model for each approach across datasets in Figure~\ref{fig:best_approach}. In addition, the comparison of first and second break-even points for the best performing models from each group are in Table~\ref{tab:break-even-points}.

\textbf{Sensitivity to the effects of randomness, and the performance variance it causes, has a significant effect on the break-even points between models, increasing the number of required labelled samples by a significant margin.} On average, fine-tuning requires $2 - 3$ times more labelled samples (increase of $100-200\%$) to outperform the remaining approaches when taking the variance into consideration. In specific cases, the impact of variance is negligible, as we observe a $0\%$ increase in the number of required samples (e.g., prompting on DB Pedia dataset). At the same time, the impact of variance is significantly higher in other cases, leading to an increase of up to $500\%$, or an increase where even with the full labelled dataset the second break-even point is not achieved (e.g., in-context learning on CoLA, BoolQ or SNIPS datasets).

\textbf{The impact of the performance variance strongly depends on the dataset and the overall number of labelled samples required for the first break-even point.} Looking at the increase in absolute numbers, an increase of $100-200\%$ represents the need to annotate only between $10 - 600$ more samples in some cases (e.g., most multi-class datasets), while representing an increase of $2000 - 11000$ or more labelled samples in other cases (e.g., binary datasets, mainly SST2). However, the \textbf{impact is most significant in cases where the second break-even point does not exist}, but can be obscured. For example on the SNIPS datasets, the first break-even point with in-context learning is achieved at $50$ labelled samples, but the second is not observed even when using all the $15 000$ labelled samples of the dataset (corresponding to an increase larger than $3000\%$).

\textbf{The second break-even point heavily depends on the variance of the approaches and models used.} In case of prompting, which shows the lowest performance variance, the highest observed increase is $157\%$ (SST2 dataset) in relative numbers or $1750$ (BoolQ dataset) in absolute numbers. On the other hand, the increase is significantly higher for in-context learning approaches (which show the highest variance across the runs) or specific cases in the instruction-tuning approaches (where the variance is more dependent on the model, with larger models showing higher variance), with the highest observed increase of $500\%$ (DB Pedia dataset and instruction-tuning) in relative numbers or increase of $11 000$ labelled samples (SST2 dataset and in-context learning) in absolute number. \textbf{The variance in the fine-tuning approaches does not have as much of an effect, as it is strongly affected by the number of labelled samples} -- the variance is higher with low number of labelled samples and almost non-existent (lower than prompting) with majority of the dataset (more than $\sim60\%$ of the labelled samples in the dataset).

\subsection{Additional Analysis: Effects of Model Size and Quantisation}

\begin{figure*}[t]
    \centering
    \includegraphics[width=0.99\textwidth]{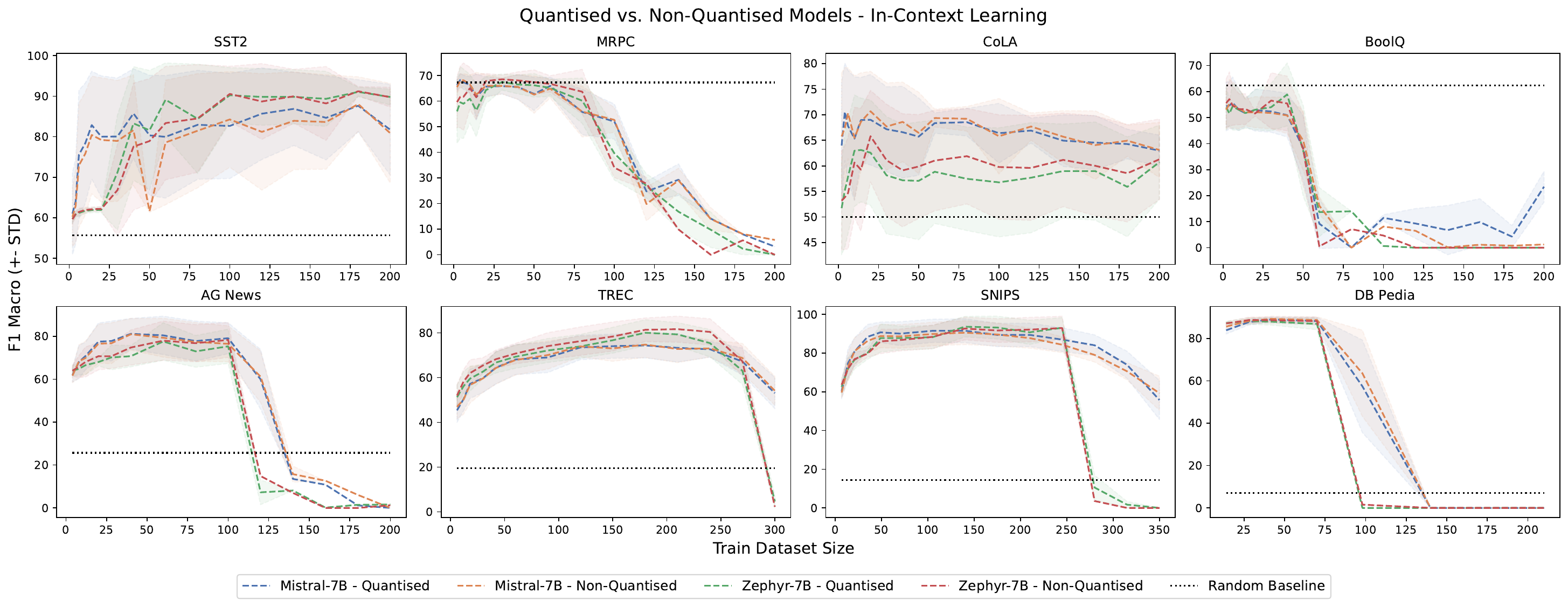}
    \caption{The comparison between 4-bit quantised and non-quantised Mistral-7B and Zephyr-7B models used for in-context learning across all datasets. The impact of quantisation is not consistent across datasets. The quantised models often achieve better performance than non-quantised ones, with the difference being often small. In addition, the impact on the variance is negligible.}
    \label{fig:quant_icl}
\end{figure*}

In this section, we provide a high-level summary of findings from analysing how the size of the model and the quantisation affect the comparison in terms of overall performance and variance. 

\textbf{Larger models do not consistently lead to better performance or lower performance variance.} The impact of model size is most explicit for fine-tuning, considerably increasing the performance and reducing variance. For prompting and in-context learning, we often observe smaller models outperforming their larger counterparts in both the performance and its variance. At the same time, larger general models often benefit more from using larger number of in-context examples. However, this is mainly effect of their larger context size (i.e., can handle more samples before observing drop in performance) and not size of the model. In addition, for prompting and in-context learning, we do not observe that using larger models leads to lower performance variance. For more details see Appendix~\ref{sec:model_size_effect}.

\textbf{Using 4-bit quantisation has minimal impact on the overall sensitivity and performance of the general models.} The quantised versions of the models often outperform their full-precision counterparts when used for in-context learning (as can be seen in Figure~\ref{fig:quant_icl}). On the other hand, for prompting, the full-precision models perform slightly better, even though the difference in performance is minimal in almost all cases. Finally, 4-bit quantised models show the same level of performance variance and benefit the same from using more in-context examples as the full-precision models. In addition, we observe negligible impact of quantisation on the performance variance. Finally, quantisation has not impact on how much the models benefit from increasing the number of in-context examples or their context length. Therefore, we can conclude that the use of quantised models has minimal impact on the comparison between different approaches. As such, considering compute-efficiency and performance trade-off in the comparison, the quantised models should be preferred for the general large language models whenever possible. For more details, see Appendix~\ref{sec:quantisation_effect}.

\section{Discussion: Recommendations Based on the Findings}
\label{section:recommendations}

In this section, we provide recommendations and suggestions that should allow for better decision making regarding what approach would provide the most efficient solution for different setting. The recommendations are based on the findings of our experiments, taking into consideration the data-efficiency (or annotation budget required for using the effective use), compute-efficiency (or the number of parameters and how it affects the whole training and inference process), the sensitivity to the effects of randomness (or how the results change when introducing small perturbation to the input data and parameters) and task characteristics.

\textbf{Using general large language models in a zero or few shot setting is a preferable solution only in specific cases}. First, when the task \textbf{does not have well defined classes} or \textbf{requires some kind of generation} (e.g., translation, summarisation). The general models excel at generation tasks (as they are designed for them) and further tuning would require an extensive dataset. Prompting is a more efficient option as providing in-context examples often requires significant human effort (e.g., preparing longer texts instead of a single label). Second, when \textbf{faced with a limited annotation and computation budget}, we do not have a large enough dataset to achieve superior fine-tuning or we do not have enough computational resources to instruction-tune the general models. However, the inference cost may be problematic in this case. Third, when we are \textbf{interested in quick prototyping and approximation of the overall performance}, before deciding whether to dedicate more time and budget for specialising the models. This recommendation is based on our main finding (specialised models require only small number of labelled samples to outperform general models) and is \textbf{in contrast to the common practice of evaluating prompting and in-context learning on classification tasks}~\citep{liu2023pretrain-prompt-survey}.

\textbf{Fine-tuning is preferable if we have large annotation budget, but small computation budget}, or if faced with task not well designed for generation, requiring additional modifications (e.g., multi-label classification). With enough samples, even small models (BERT) can outperform all the general models. Training the largest possible model (allowed by the computation budget) should be opted for, to reduce the number of required samples as much as possible, and to prevent any shortcomings on tasks that require better language understanding. Based on our findings, we \textbf{need to annotate on average $100$ samples, even for smaller models.} Obtaining such a number of labelled samples is not problematic in many domains.

\textbf{Instruction-tuning of the larger language models is the optimal solution as it provides consistent benefits for all the other cases} (large computation budget with either small or large annotation budget). Instruction tuning provides the best trade-off for the performance and the required resources -- increasing the number of labelled samples does not have a significant impact on the performance, while increasing the size of the instruction-tuned model has only low performance impact. However, if interested in best overall performance on the task, using the largest general model and using as many labelled samples for instruction-tuning (and in-context examples) should provide the most benefit at the cost of increase in effort (and strong diminishing returns). In addition, it requires significantly larger computation budget than fine-tuning, due to the inference costs, which should be considered in the long run.

\textbf{Using 4-bit quantisation can lead to higher compute-efficiency with minimal impact on the performance.} The impact of quantisation on the performance in text classification is minimal (achieving similar or higher performance). The impact on sensitivity is negligible. As such, the trade-off between significantly lower training/inference costs and similar performance favours the quantised models for prompting, in-context learning and instruction-tuning.

\textbf{When comparing between different approaches and models, the performance variance should be taken into consideration}, as it has a strong impact on the comparison. If only a single run (or low number of runs) is used, one of the models can be incorrectly denoted as better only based on a random chance. The use of multiple runs is especially important when using in-context learning, instruction-tuning or fine-tuning on low number of labelled samples. 

Finally, when using general models, \textbf{a specific focus should be on the prompt format and the set of in-context examples to maximise their benefits.} The prompt format was identified as the most significant contributor to the performance and variance in the results. To achieve fair comparison and robustness of the models, the comparison should be done using prompt optimised for each model or an ensemble of multiple prompts for instruction tuning and evaluation. In addition, \textbf{choosing the optimal number of in-context examples is important to achieve fair comparison.} Using all the available samples may not always be possible or beneficial due to the limited context size. As illustrated in our experiments, the performance drops significantly after this limit is reached (or sometimes even beforehand). Furthermore, \textbf{the quality of the samples is as important (if not more) than the quantity of the samples.} Majority of the observed variance in the performance in our experiments is the result of the random choice of examples. Previous studies have also observed that the quality of the samples is paramount, even though identifying the samples of "highest quality" may not be as straightforward~\citep{pecher2024automatic, agrawal-etal-2023-context, chang-jia-2023-data}. \textbf{As such, both quality and quantity of the in-context examples should be considered in the comparisons, introducing further human labour that should be considered when choosing the most effective and efficient approach and model.}

\section{Conclusion}

In this paper, we perform a comprehensive and fair comparison of currently popular approaches for data-efficient learning in NLP, from the perspective of their data-efficiency and sensitivity to the effects of randomness. The main focus of the investigation is to determine how many labelled training samples are needed for the specialised small language models (obtained through fine-tuning or instruction-tuning) to outperform their general larger counterparts used in zero and few-shot settings. Based on the break-even points from 8 representative text classification datasets of various characteristics, we find that the number of required training samples is quite small, but strongly dependent on the dataset and task characteristics, with more labelled samples needed for binary datasets and tasks that require better language understanding. In addition, we identify a significant influence of the performance variance, stemming from the sensitivity of the different approaches, especially in-context learning and fine-tuning with few labelled samples, on the overall comparison and the number of required labelled samples. We can conclude that the specialised smaller models are a strong contender on the text classification tasks, with the general language models showing their benefits for specific cases, such as quick prototyping or when faced with extremely limited annotation. Finally, based on our findings, we provide recommendations that allow for better decision making regarding what approach to use for different settings.

\section*{Acknowledgements}
This work was partially supported by the projects funded by the European Union under the Horizon Europe: \textit{DisAI}, GA No. \href{https://doi.org/10.3030/101079164}{101079164}, \textit{AI-CODE}, GA No. \href{https://cordis.europa.eu/project/id/101135437}{101135437}; by \textit{Modermed}, a project funded by the Slovak Research and Development Agency, GA No. APVV-22-0414; and by the European Union NextGenerationEU through the Recovery and Resilience Plan for Slovakia under the project No. 09I03-03-V03-00020.

Part of the research results was obtained using the computational resources procured in the national project \textit{National competence centre for high performance computing} (project code: 311070AKF2) funded by European Regional Development Fund, EU Structural Funds Informatization of Society, Operational Program Integrated Infrastructure.

\section*{Limitations}

The investigation is done on a set of English classification datasets with various characteristics (number of classes, input size, task type, etc.). This choice may limit the generalisation of our findings to other datasets, tasks and languages, such as the generation tasks, which are more representative for the general models and on which the fine-tuning cannot be used as easily (e.g., question answering, summarisation, translation). However, we explicitly discuss this in Section~\ref{section:recommendations}.

Due to the significant computation costs incurred by the inference of larger language models (namely LLaMA-2, LLaMA-3, Mistral-7B and Zephyr-7B), instruction-tuning of the medium sized models (Mistral/Zephyr) and the additional other costs of the GPT-4 model, we evaluate the models in a limited setting. The evaluation is done on a randomly selected set of 1 000 samples (following the practice in prior works~\cite{gao-etal-2021-making, chang-jia-2023-data, sclar2024quantifying, li-qiu-2023-finding, koksal-etal-2023-meal}). As the datasets are smaller (e.g., MRPC, BoolQ, CoLA, SNIPS, TREC) in many cases, this decision may not be as problematic, as the used 60:20:20 split leads to approximately the 1000 test samples. However, on the larger datasets (SST2, AG News and DB Pedia) this decision may skew the results. In addition, we run the GPT-4 prompting and in-context learning only 10 times instead of 20 due to its costs.

We use the same prompt for all the models, which is a result of a prompt engineering on the Mistral-7B model. The prompt was created based on dataset description, the prompts used in related works and the formats recommended for different models (e.g., taking inspiration from~\cite{sun2023pushing}). As such, this may not represent the optimal format for all the models (as identified in previous works~\cite{sclar_quantifying_2023, weber-etal-2023-mind, pecher2023effects}) and performing the investigation on multiple different prompts may improve the overall model performance and affect the findings. However, we opted for using only a single optimised prompt format to reduce the computation costs and provide more in-depth analysis on larger number of models.

Finally, we are not sure whether the datasets we use in our experiments have been used to pre-train the models we use for prompting and in-context learning. As such, the comparison and findings of our study may be affected by this possible data leak, leading to larger benefit of general models over the specialised ones (as already indicated in Section~\ref{sec:first-bep}). We limit this effect by using a diverse set of datasets and our own optimised prompt across the majority of the experiments. However, we cannot guarantee it is enough to provide unbiased results as this limitation is part of the recently recognised LLM validation crisis~\citep{li2023task} and we would need to train the model from scratch to address it properly, which is out of scope for this paper.

\section*{Ethical Considerations}
The experiments in this paper work with publicly available benchmark datasets GLUE and SuperGLUE and other publicly available datasets (AG News, TREC, SNIPS and DB Pedia), citing the original authors. As we were not able to determine the license for the tasks used, we opted to use them in as limited form as possible, adhering to the terms of use (no annotation of the test set) defined by the GLUE and SuperGLUE and applying it to other datasets as well. We do not work with any personally identifiable information or offensive content and perform no crowdsourcing for further data annotation. In addition, we are not aware of any potential ethical harms or negative societal impacts of our work, apart from the ones related to the advancement of the field of machine learning. Finally, we follow the license terms for all the models we use (such as the one required for the use of the LLaMA-2 and LLaMA-3 models). It is possible the large language models we use contain biases and potentially offensive or harmful content. However, the original authors of these models reduce this bias as much as possible. At the same time, we do not release any output of the models which should further reduce the potential bias and negative impact.

\bibliography{custom_bib}

\appendix

\section{Experimental Setup: Additional Details}
\label{app:experimental-setting-details}

Each experiment is repeated 100 times for fine-tuning and 20 times for the remaining approaches (due to the significant costs of inference or training of the larger language models) in order to observe and reduce the performance variance (as recommended by~\citeauthor{gundersen2023reporting}~\cite{gundersen2023reporting} and \citeauthor{pecher2023effects}~\cite{pecher2023effects}), since particularly in-context learning and fine-tuning with few samples were identified to produce results with significant variance due to effects of randomness~\cite{lu-etal-2022-fantastically, zhao2021calibrate, zhang-etal-2022-active, mosbach_stability_2021, dodge2020fine}. Each repeat uses the same fixed random seed for all models to guarantee deterministic and replicable results. The random seed covers all sources of randomness~\cite{gundersen2022sources} -- the split of data, selection of labelled samples, initialisation of models, order of data in training, non-deterministic operations (e.g., dropout), and selection of samples for in-context learning. At each point, we report the mean F1 macro and standard deviation. 

\subsection{Dataset Details}

All experiments in this paper use English-only datasets. We focus on 8 datasets composed of diverse tasks with different number of classes and characteristics. We focus on 4 binary datasets from GLUE~\cite{wang-etal-2018-glue} and SuperGLUE~\cite{wang2019superglue} benchmarks: \textit{SST2}~\cite{socher-etal-2013-recursive} for sentiment classification, \textit{MRPC}~\cite{dolan-brockett-2005-automatically} for determining semantic equivalence relationship between two sentences, \textit{CoLA}~\cite{warstadt-etal-2019-neural} for determining the grammatical acceptability of a sentence, and \textit{BoolQ}~\cite{clark-etal-2019-boolq} for question answering. In addition, we use 4 multi-class text datasets: \textit{AG News}~\cite{zhang2015agnews} for news classification (4 classes), \textit{TREC}~\cite{voorhees2000trec} for question classification (6 classes), \textit{SNIPS}~\cite{coucke2018snips} for intent classification (7 classes) and \textit{DB Pedia}~\cite{lehmann2015dbpedia} for topic classification (14 classes).

For each dataset, we split all the available labelled samples into training, validation and test set using 60/20/20 split. In addition, we subsample each dataset, starting with 1 labelled sample per class and then increasing this number exponentially up to the full dataset (or a dataset fraction where fine-tuning outperforms all other approaches). Both the data split and subsampling is determined by the random seed. We choose this setup (i.e., creating new split of samples instead of using the pre-defined dataset splits), as some of the datasets do not release labels for test sets and the validation splits contain quite low number of sample, possibly leading to unreliable results. In addition, this guarantees that the setup is as similar as possible across all the datasets.

\subsection{Approaches and Models Setup}

\begin{table}[tbh]
\begin{center}
\tabcolsep=0.08cm
\begin{tabularx}{\linewidth}{@{}p{0.15\linewidth}p{0.825\linewidth}@{}}
\toprule
\multicolumn{2}{X}{\textbf{Prompt Format}}  \\ \midrule
\multicolumn{2}{X}{Determine \{ sentiment | grammatical acceptability | semantic equivalence | whether the passage contains answer | topic | intent \} of the \{ sentence | sentence pair | question \} using following options: 1) \textit{[Class 1]} 2) \textit{[Class 2]} ... N) \textit{[Class N]}. \vspace{2mm} \newline \textit{[Input]} \newline \textit{[Output]}} \\ \midrule \midrule
\textbf{Dataset} & \textbf{Verbaliser}  \\ \midrule
\textit{SST2}       & \{Negative, Positive\} \\
\textit{CoLA}        & \{No, Yes\} \\
\textit{MRPC}        & \{No, Yes\} \\
\textit{BoolQ}       & \{No, Yes\} \\
\textit{AG News }    & \{World, Sports, Business, Science and Technology\} \\
\textit{TREC}        & \{Expression, Entity, Description, Human, Location, Number\} \\
\textit{SNIPS}       & \{Playlist, Weather, Event, Musing, Creative Work, Rate Book, Book Restaurant\} \\
\textit{DB Pedia}    & \{Company, Educational Institution, Artist, Athlete, Office Holder, Transportation, Building, Natural Place, Village, Animal, Plant, Album, Film, Written Work\} \\
\bottomrule
\end{tabularx}
\end{center}
\caption{Prompt formats and verbalisers used for prompting, in-context learning and instruction-tuning across different datasets. The \textit{[Class 1-N]} and the \textit{[Output]} are replaced with the names of the classes defined by the verbaliser. The \textit{[Input]} is replaced by the sentence of the samples. The \textit{[Input]} and \textit{[Output]} are repeated for each in-context sample, while the final \textit{[Output]} is used to determine the predicted class.}
\label{tab:prompt-format}
\end{table}

For fine-tuning, we use the \textbf{BERT}~\citep{devlin-etal-2019-bert} and \textbf{RoBERTa}~\citep{liu2019roberta} base models. We follow the typical setup and recommendations from related work (such as~\citet{mosbach_stability_2021, dodge2020fine}), adding dropout layer with drop rate of 0.3 followed by a classification layer. We use learning rate of 1e-5 with AdamW optimiser with warmup and train the model until convergence using a maximum of 10 epochs, with variable batch size across different dataset subsets (starting at 4 and ending at 32 for the full dataset). 

For prompting and in-context learning the \textbf{Flan-T5}~\citep{chung2024scaling} base, \textbf{GPT-4} (4o-mini-2024-07-18 version)~\citep{brown2020language, ouyang2022training}, \textbf{LLaMA-2}~\citep{touvron2023llama} 13B chat optimised model, \textbf{LLaMA-3}~\citep{dubey2024llama} 8B instruction optimised model, \textbf{Mistral-7B}~\citep{jiang2023mistral} and \textbf{Zephyr-7B}~\citep{tunstall2023zephyr} models. For instruction-tuning we use the \textbf{Flan-T5} (with full instruction-tuning), \textbf{Mistral-7B} and \textbf{Zephyr-7B} models (both instruction-tuned using LoRA~\citep{hu2022lora}).

The prompt format used for prompting, in-context learning and instruction tuning is included in Table~\ref{tab:prompt-format}. It is a result of a simple prompt engineering based on prompt formats used in related work and their recommendations~\citep{sun2023pushing}.

For in-context learning, we use as many samples per class as the context-size of the models allow. As increasing this number after a certain point results in degraded performance, we always report the best performance achieved up until the dataset subset. As such, in some cases, we report the performance using 2-shot in-context learning classification even on full datasets, while in others, we report 100-shot. This is often the case with GPT-4 model as it does not appear to benefit from increasing the number of in-context examples as much.

Instruction tuning is done using the HuggingFace SFT trainer, with a learning rate of 1e-5 for 5 epochs using AdamW optimiser with warmup, batch size of 4 and early stopping, with a maximum of 250 steps per epoch for the Flan-T5 model (as we observed severe overfitting of Flan-T5 without this constraint on larger dataset sizes) and no maximum steps for the remaining models. For this tuning, we preprocess the dataset using the data collator in order to train only on the completions (as described in the documentation\footnote{\url{https://huggingface.co/docs/trl/en/sft_trainer}}).

All of the general models (Flan-T5, LLaMA-2, LLaMA-3, GPT-4, Mistral-7B and Zephyr-7B) are set to be deterministic (e.g., using no sampling, no beam search and using deterministic temperature) and set maximum number of tokens for generation to 10. In addition, due to the significant costs, we use 4-bit quantisation for the LLaMA-2 model. In case of Mistral-7B and Zephyr-7B, we use both the non-quantised version and the 4-bit quantised versions. For instruction-tuning we use only the 4-bit quantised versions of Mistral-7B and Zephyr-7B models due to the significant cost of training the models. The Flan-T5 model, the LLaMA-3 model and the GPT-4 model are always used with full-precision.

Finally, for better results and comparison presentation, we group the individual models by size. We define the following 3 sizes: 1) \textit{small}, which includes the Flan-T5 model; 2) \textit{medium}, which includes the Mistral-7B and Zephyr-7B models; and 3) \textit{large}, which includes the LLaMA-2, LLaMA-3 and GPT-4 models.

\paragraph{Using subsets of different sizes.} The way the different subsets of labelled training samples are used is dependent on the approach for learning with limited labelled data. In case of prompting, no training samples are utilised and as such the performance remains the same across regardless of the dataset size. In case of in-context learning, either all of the samples are used as in-context examples or, in case when the context-size of the model is not enough to use all samples, a subset of samples is randomly chosen. In case of fine-tuning, all the labelled samples are used for training. Finally, for instruction-tuning we combine the fine-tuning and prompting/in-context learning methodology, i.e., all of the samples are first used to tune the model and afterwards it is used through prompting and in-context learning the same as the models without parameter updates.

\paragraph{In-context learning results reported for the main experiments.} As we observe that the 4-bit quantised models outperform the full-precision versions on some dataset, to make the results more understandable, we report them only for the best performing models. The results are reported in the following way (which can be determined from Figure~\ref{fig:quant_icl}):
\begin{itemize}
    \item For in-context learning we report the results from 4-bit quantised Mistral and Zephyr on the SST2, BoolQ, and SNIPS datasets; from the 4-bit quantised Mistral on the MRPC and AG News datasets; and from the full-precision version for all the remaining combination.
    \item For prompting, we report the results from the 4-bit quantised version of Mistral and Zephyr only for the SST2 and DB Pedia datasets; from the 4-bit quantised Mistral for AG News datasets; from the 4-bit quantised Zephyr for the MRPC dataset; and from the full-precision version for all the remaining combinations.
    
\end{itemize}

\subsection{Model Computation Requirements}
\label{app:computation-flops}

To allow for a fair comparison, the models and approaches used in this study were specifically chosen to require comparable computation for their whole use. For example, the full fine-tuning and evaluation of the smallest model (BERT) takes approximately the same amount of compute as using the smallest LLM (Flan-T5) through prompting. Similarly, instruction-tuning and further evaluation with the smallest LLM takes approximately the same amount of compute as using the larger LLMs through in-context learning. However, it is important to note that this holds true only when using a low number of shots for in-context learning (up to 2). Using more in-context examples leads to significantly larger computation costs. In Table~\ref{tab:flops} we provide an approximation of FLOPs for the approaches we use following the methodology from related papers, such as~\citet{brown2020language, dubey2024llama}.

\begin{table*}[]
\begin{center}
\begin{tabular}{cr}
\toprule
\textbf{Model}                                   & \textbf{Approximate FLOPs} \\ \midrule
Fine-Tuning (BERT, RoBERTa)                      & 270 T                      \\
Small general LM (Flan-T5) - Prompting           & 300 T                      \\
Medium general LM (Mistral/Zephyr) - Prompting   & 710 T                      \\
Large general LM (LLaMA2) - Prompting            & 1285 T                     \\
Small general LM (Flan-T5) - ICL                 & 804 T                      \\
Medium general LM (Mistral/Zephyr) - ICL         & 1900 T                     \\
Large general LM (LLaMA2) - ICL                  & 3400 T                     \\
Instruction-Tuning Small (Flan-T5)               & 1200 T                     \\
Instruction-Tuning Medium (Mistral/Zephyr)       & 2800 T                     \\
Quantised medium LM (Mistral/Zephyr) - Prompting & 710 T (308 T)              \\
Quantised medium LM (Mistral/Zephyr) - ICL       & 1900 T (826 T)             \\
\bottomrule
\end{tabular}
\end{center}
\caption{Approximate FLOPs for different models and approaches we use that are required for their full use. As full use, we understand the full tuning and evaluation in the case of fine-tuning and instruction-tuning or only the evaluation in the case of prompting and in-context learning. For the quantised models, we report the effective FLOPs in parenthesis that are normalised by the bit-width.}
\label{tab:flops}
\end{table*}

\section{Effect of Model Size on Comparison and Dataset Dependence}
\label{sec:model_size_effect}

\begin{figure*}[t]
    \centering
    \includegraphics[width=0.99\textwidth]{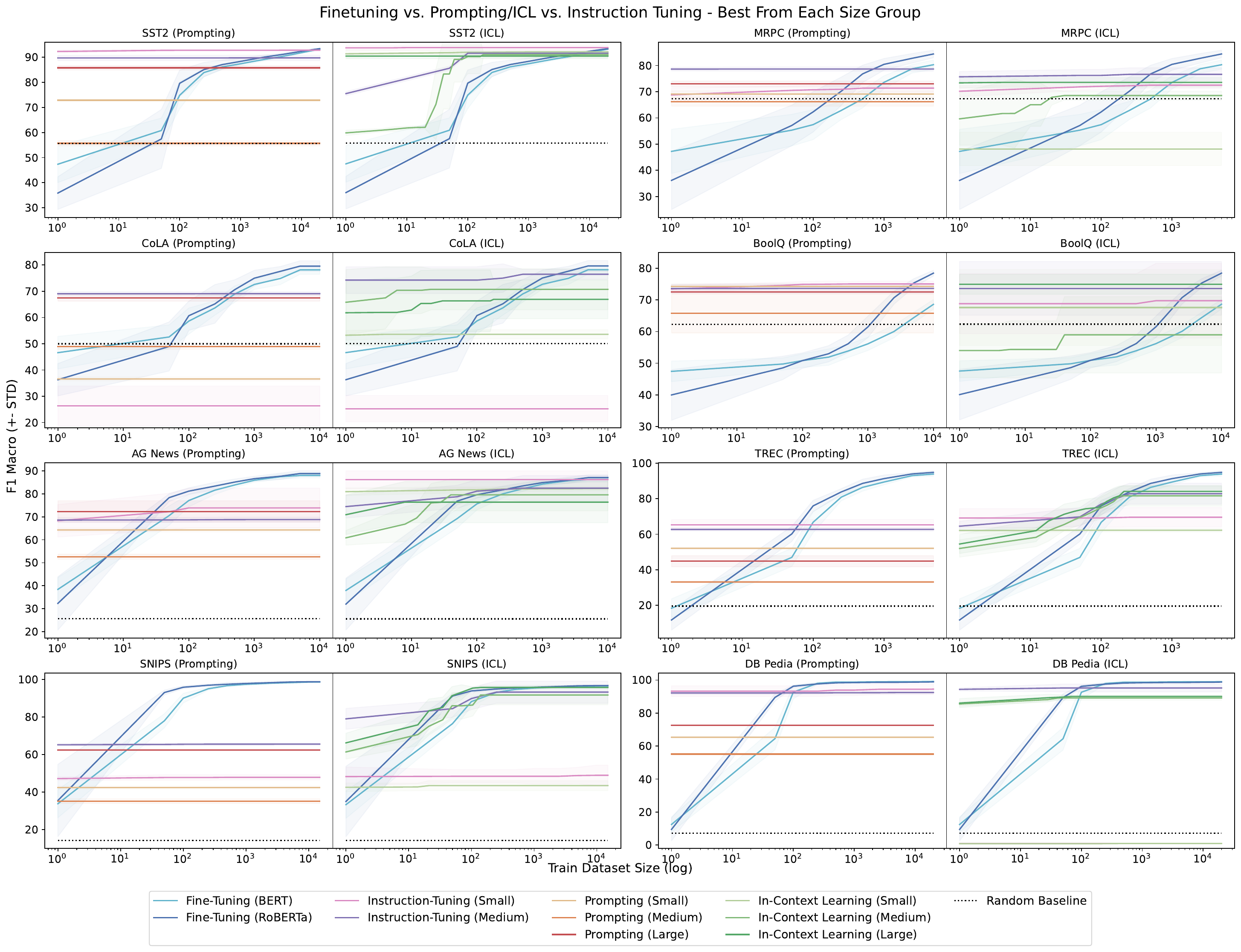}
    \caption{The impact of varying size of available labelled training samples (in logarithmic scale) on the performance of fine-tuning, prompting, in-context learning and instruction-tuning approaches across the binary (SST2, MRPC, CoLA, BoolQ) and multi-class datasets (AG News, TREC, SNIPS, DB Pedia), reported using F1 macro and its standard deviation. For each approach, we group the models based on their size into small (Flan-T5), medium (Mistral/Zephyr) and large (LLaMA/GPT) and report the best performing model for each group. Even though the effect of model size is often significant, it does not follow common assumption. \textbf{The smaller models, especially in prompting, in-context learning and instruction-tuning, often achieve better performance than the medium or large general models.}}
    \label{fig:group_best_approach}
\end{figure*}

\begin{figure*}[tbh]
    \centering
    \includegraphics[width=0.99\textwidth]{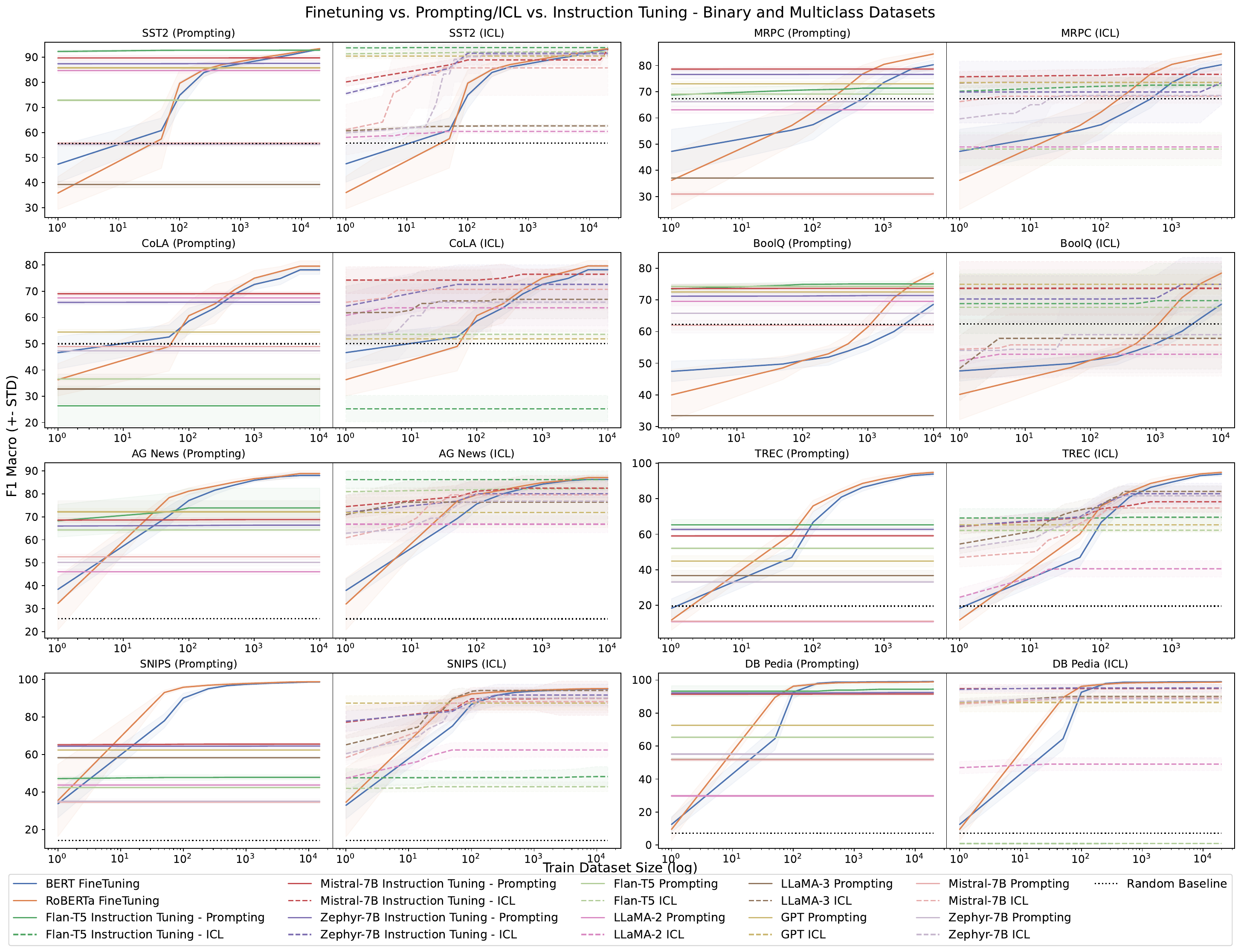}
    \caption{The impact of varying size of available labelled training samples (in logarithmic scale) on the performance of fine-tuning, prompting, in-context learning and instruction-tuning approaches, reported using F1 macro and its standard deviation. All the models are presented for each approach.}
    \label{fig:meta-full-image}
\end{figure*}

Previous studies have observed that the size of the model significantly affects its ability to perform prompting and in-context learning. As such, in this section, we are interested in how the size of the model and other factors affect the comparisons, both in term of overall performance and variance, but also other properties as well. Instead of reporting results only for the best performing model for each approach, we group the models based on their size into small (Flan-T5), medium (Mistral/Zephyr) and large (LLaMA/GPT) and report the best performing model for each group. The results are included in Figure~\ref{fig:group_best_approach}. In addition, we provide the results for the individual models for each approach and dataset in Figure~\ref{fig:meta-full-image}.

First of all, \textbf{larger models do not consistently lead to better performance} across different approaches. With the fine-tuning, we observe the expected behaviour. On lower number of labelled samples (up to $20 - 100$ depending on the dataset) the smaller BERT model is able to outperform the larger RoBERTa model. However, the larger model can achieve the highest performance on the different datasets using fewer labelled samples. As such, the individual break-even points appear earlier for the larger fine-tuning models. On the other hand, in case of prompting, in-context learning and instruction-tuning, the smaller models are often able to achieve better or similar performance than the larger counterparts even though the different in the number of parameters is large. For example, in-context learning with small models (Flan-T5) on the AG News and SST2 dataset outperforms in-context learning with medium (Mistral-7B and Zephyr-7B) or large (LLaMA and GPT) models, while on the BoolQ dataset the situation is opposite. Similarly, instruction-tuning with medium models (Mistral-7B and Zephyr-7B) models often leads to similar or lower performance than the instruction-tuning of the small (Flan-T5) model. 

\textbf{This behaviour does not necessarily depend on the dataset characteristics.} On some datasets that can be considered harder (longer inputs, more classes, harder tasks that require better language understanding), the Flan-T5 model in zero/few shot setting still outperforms the larger models, while on other datasets it underperforms them. The main impact of dataset characteristics is causing a 'failure mode' for specific models, where the smaller model does not perform at all at the task. For example, using in-context learning with small (Flan-T5) model on the DB-Pedia dataset always leads to prediction of a single class for all of the samples, with this behaviour disappearing when reducing the number of classes, or using prompting. Similarly, in-context learning with the smaller (Flan-T5) model on the BoolQ dataset (which is characteristic with its longer input) shows a significantly higher variance. We believe the main culprit for this is the limited context size of the smaller model.

\textbf{The impact of the model size is the most explicit for fine-tuning.} On specific datasets that can be considered harder, the smaller BERT model performs significantly worse. For example, the BERT model on the BoolQ fails to outperform majority of the remaining approaches even when using the full dataset. However, this impact is minimal on the multi-class datasets.

Overall the \textbf{number of required labelled samples is not consistently dependent on the size of the general models.} For prompting and in-context learning, we observe many cases when the break-even point for the smaller model requires similar or even larger number of labelled samples than the medium or large models. For example, on the AG News dataset the number of required labelled samples for RoBERTa to outperform Flan-T5 in-context learning is $300$ (or $800$ when taking the variance into consideration), while for the medium models it is only $100$ and for large models it is $50$. Similar observations are present in instruction tuning as well. Curiously, fine-tuning can often outperform the medium sized models (Mistral/Zephyr) with lower number of samples than the smallest Flan-T5 model. In addition, on the binary datasets, fine-tuning consistently requires the highest number of samples to outperform the largest models (LLaMA and GPT).

\textbf{The benefit of increasing the number of in-context examples for in-context learning is affected by the size of the models as well.} However, this benefit is, again, dataset dependent and is closely tied to the context length and the dataset characteristics. For example, on the binary datasets with shorter sentences (SST2/CoLA), the benefit of more samples is significantly higher than on the datasets with multiple classes or with longer sentences (BoolQ/DB Pedia). This also directly affects the benefit of instruction-tuning. Although the benefit is larger for the medium and large models, in many cases increasing the number of in-context samples only leads to these models to achieve performance similar to the one achieved by the smaller model -- i.e., for in-context learning with medium models, we require up to $100$ samples to achieve performance comparable to small models that use only 10 labelled samples.

In addition, we \textbf{do not observe consistent improvement of in-context learning (few-shot setting) over prompting (zero-shot setting).} Instead, the performance difference is more dataset dependent. On the multi-class datasets, in-context learning is beneficial over prompting in majority of the cases. On the binary datasets, the benefit of in-context learning strongly depends on the models. At the same time, the improvement of in-context learning over prompting is minor in some cases (such as the increase in Flan-T5 on the SNIPS dataset from $42.5\%$ to $44.5\%$).

Finally, \textbf{using larger models does not necessarily lead to lower performance variance.} With the fine-tuning models, we observe the expected behaviour again. Larger model shows lower variance when using enough labelled data, but shows larger variance than the smaller model when dealing with limited data. The performance variance in prompting is similar across all the models, as there are almost no sources of randomness that could affect it. On the other hand, the variance in in-context learning is not as consistent. Comparing the smallest model (Flan-T5) and the largest models (LLaMA and GPT), we observe similar performance variance on majority of the datasets. Only on specific datasets the variance of Flan-T5 is significantly higher (e.g., BoolQ) or significantly lower (AG News). At the same time, the variance of the medium sized models (Mistral-7B and Zephyr-7B) is often significantly higher than their smaller or larger counterparts. Finally, with instruction-tuning, we observe that the medium models (Mistral-7B and Zephyr-7B) often show larger variance when compared to Flan-T5. Although this may be due to the different type of instruction-tuning (full tuning in Flan-T5 vs. LoRA tuning in Mistral/Zephyr), we observe similar behaviour of these models in the zero/few shot setting as well.

\section{Effect of Quantisation on General Models}
\label{sec:quantisation_effect}

\begin{figure*}[t]
    \centering
    \includegraphics[width=0.99\textwidth]{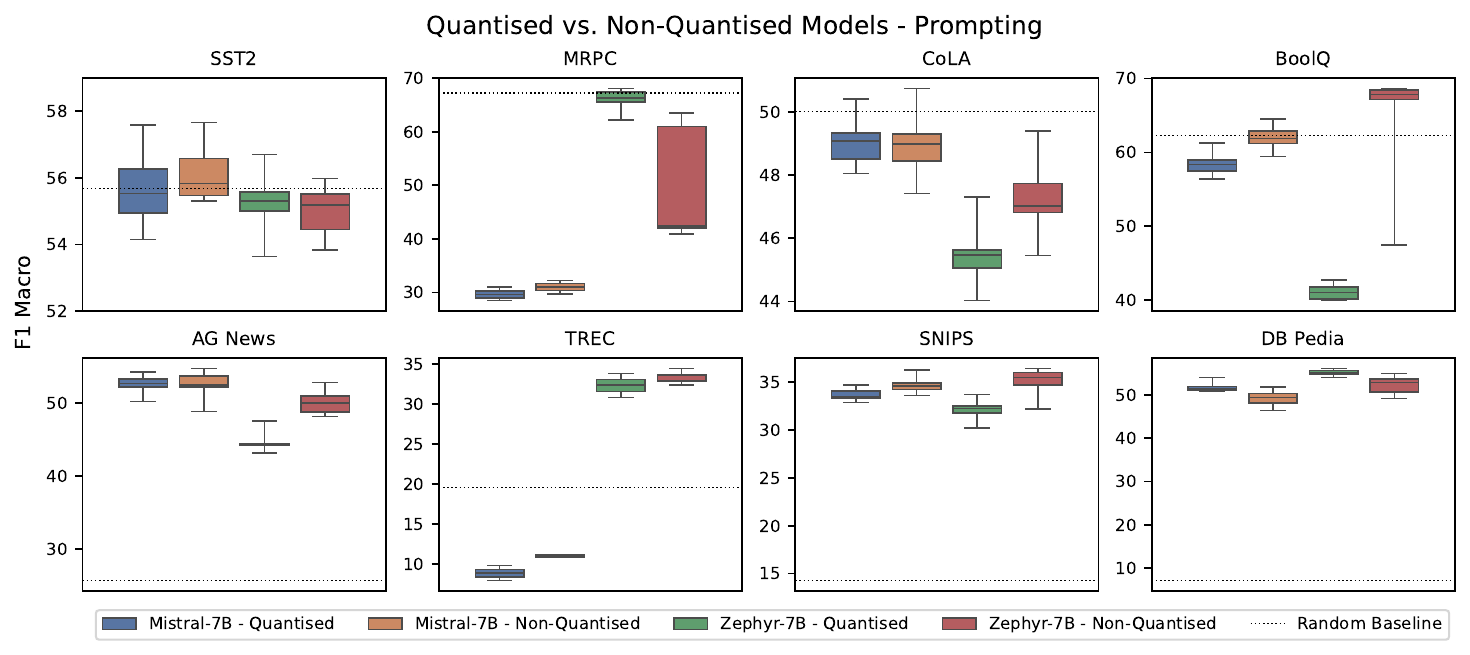}
    \caption{The comparison between 4-bit quantised and non-quantised Mistral-7B and Zephyr-7B models used for prompting across all datasets. Even though the non-quantised models achieve better performance, the difference in performance is minimal in almost all cases. In addition, the impact on the variance is negligible.}
    \label{fig:quant_prompting}
\end{figure*}


As the general large language models contain high number of parameters, it is a common practice to use their quantised versions instead to reduce the computation costs. Although some paper research focuses on observing how this affects the performance, the impact of quantisation on the comparison between models and the sensitivity to the effects of randomness is not well understood. Therefore, in this section, our aim is to determine how the use of quantised models affect the comparison and the overall sensitivity to the effects of randomness. To achieve this, we run the same experiment with 4-bit quantised and non-quantised versions of the Mistral-7B and Zephyr-7B models using the prompting and in-context learning and compare their average performance and variance. The results of this comparison are presented in Figure~\ref{fig:quant_prompting} for prompting and in Figure~\ref{fig:quant_icl} for in-context learning.

\textbf{The non-quantised models perform slightly better than 4-bit quantised models when it comes to prompting.} The difference in performance between the 4-bit quantised and non-quantised models is minimal in almost all the cases, with non-quantised models achieving slightly higher performance. Only exception is the MRPC dataset, where the Zephyr 4-bit quantised model achieves better performance, and DB Pedia where both Mistral and Zephyr achieve higher performance. On the other hand, the non-quantised version of the Zephyr model achieves significantly higher performance on the BoolQ, CoLA and AG News datasets.

The impact of quantisation is less consistent for in-context learning. In contrast to prompting, \textbf{the 4-bit quantised models used for in-context learning outperform their non-quantised versions more often.} In addition, the difference in performance in such cases is often significantly higher. At the same time, when the quantised models do not outperform the non-quantised versions, the difference in performance is often small (only exception is the Zephyr model on the CoLA dataset).

Furthermore, \textbf{the impact of quantisation on the sensitivity to the effects of randomness and the performance variance it causes is negligible.} In almost all cases, the observed standard deviation is similar regardless of the model version. In case the performance variance is different, the non-quantised models achieve higher performance. For example, the prompting of Zephyr model on the MRPC, BoolQ or AG News datasets. In case of in-context learning, the difference in performance variance is even less significant.

Finally, \textbf{quantisation has no impact on how much the models benefit from increasing the number of in-context examples or their context length.} In all cases, we observe similar increase in performance (and decrease in standard deviation) when increasing the number of examples. At the same time, we observe that both quantised and non-quantised versions of the models achieve the `failure state` at the same number of samples. We believe this failure state corresponds to reaching the context size of the models, as on the datasets with longer sentences (BoolQ/DB Pedia), it is encountered sooner than on datasets with shorter sentences (SST2/CoLA). However, in many cases, \textbf{the best performance of the model is achieved well before using the whole context size of the models.}

\textbf{We can conclude that the use of quantised models has minimal impact on the comparison between different approaches.} As such, considering compute-efficiency and performance trade-off in the comparison, the quantised models should be preferred for the general large language models whenever possible.

\section{Impact Statement: CO2 Emissions Related to Experiments}

The experiments presented in this paper used significant compute resources as they required multiple training and evaluation runs of multiple models (to deal with variance in results), as well as using large language models that requires a lot of computation even just for the inference. Overall, the experiments were conducted using a private infrastructure, which has a carbon efficiency of 0.432 kgCO$_2$eq/kWh (default value used as the actual efficiency of our HW instance was not measured). A cumulative of 4000 hours of computation was performed on hardware of type A100 PCIe 40GB (TDP of 250W). Total emissions are estimated to be 432 kgCO$_2$eq of which 0 percents were directly offset. This does not include the compute used by the GPT model behind API as we are not able to estimate these resources. These estimations were conducted using the \href{https://mlco2.github.io/impact#compute}{MachineLearning Impact calculator} presented in \cite{lacoste2019quantifying}.

Whenever possible, we tried to reduce the compute resources used as much as possible. The most compute resources were used by the large language models -- LLaMA-2, LLaMA-3, GPT-4, Mistral-7B and Zephyr-7B. As the prompting and in-context learning with these models resulted in quite stable results, we decided to evaluate them only on a single setting (using 1 000 labelled training samples) and using a fraction of the whole test set (1 000 test samples). In addition, for the GPT model, we evaluate only on 10 runs. Even in their reduced evaluation, these experiments used large fraction of the GPU hours. The most significant contributor was the instruction-tuning, especially with the medium sized models (Mistral/Zephyr), where we opted to reduce the number of steps and epochs used for the training.
\label{compute-used}

\end{document}